\documentclass[11pt,letterpaper]{article}

\usepackage{naaclhlt2018}
\usepackage{times}
\usepackage{url}
\usepackage{amsmath}
\usepackage{amsthm}
\usepackage{amsfonts}
\usepackage{latexsym}
\usepackage{microtype}
\usepackage{booktabs}

\usepackage[small]{caption}
\usepackage{subcaption}

\usepackage{marginnote}

\usepackage[disable]{todonotes}
\makeatletter
\newcommand*\iftodonotes{\if@todonotes@disabled\expandafter\@secondoftwo\else\expandafter\@firstoftwo\fi}  
\makeatother

\newcommand{\note}[4][]{\todo[author=#2,color=#3,size=\scriptsize,fancyline,caption={},#1]{#4}} 
\newcommand{\ryan}[2][]{\note[#1]{ryan}{violet!40}{#2}}

\usepackage{mathtools} 
\usepackage{qtree}

\long\def\eat#1{\ignorespaces}

\usepackage{microtype}
\usepackage{cleveref}
\crefname{section}{\S}{\S\S}
\Crefname{section}{\S}{\S\S}
\crefname{table}{Table}{}
\crefname{figure}{Figure}{}
\crefname{algorithm}{Algorithm}{}
\crefname{appendix}{Appendix}{}
\crefformat{section}{\S#2#1#3}  

\aclfinalcopy 
\usepackage{algorithm}
\usepackage[noend]{algpseudocode}

\newcounter{notecounter}

\newcommand{\enoteson}{\long\gdef\enote##1##2{{

\stepcounter{notecounter}
\large\bf
\hspace{1cm}\arabic{notecounter} $<<<$ ##1: ##2 $>>>$\hspace{1cm}}}}
\enoteson

\renewcommand{\algorithmicindent}{9pt}
\algnewcommand{\LineComment}[1]{\State \(\triangleright\) {\small \it #1}}
\algnewcommand{\InlineComment}[1]{\hfill \(\triangleright\) {\small \it #1}}
\algrenewcommand\algorithmicindent{1.0em}%

\usepackage[safe]{tipa}
\usepackage{amssymb}

\def\ouremph#1{\textbf{#1}}

\title{  \raisebox{1ex}[0in][0in]{\parbox[b]{\linewidth}{\begin{flushright}\footnotesize
        \textmd{\textsf{\textcolor{gray}{Appeared in the
                proceedings of EMNLP 2016 (Austin, November).  This
          version was \\ prepared in January 2021 and is clarified and fixes a mistake in the gradient of the log-likelihood.}}}\end{flushright}}}\\ \vspace{-1ex}Morphological Segmentation Inside-Out}
          
          \author  
  {
\begin{tabular}{cccc}
	Ryan Cotterell\raise1.0ex\hbox{\normalfont\normalsize\textschwa} & Arun Kumar\raise1.0ex\hbox{\normalfont\normalsize\textipa{G}} & Hinrich Sch{\"u}tze\raise1.0ex\hbox{\normalfont\normalsize\textipa{H}}
	\end{tabular}
	\\
    \raise1.0ex\hbox{\normalsize\textschwa}Department of Computer Science, Johns Hopkins University \\
     \raise1.0ex\hbox{\normalsize\textipa{G}}Faculty of Arts and Humanities, Universitat Oberta de Catalunya \\
    \raise1.0ex\hbox{\normalsize\textipa{H}}CIS, LMU Munich  \\
	{\tt\small{\{ryan.cotterell\}@jhu.edu}}
}

\usepackage{qtree}
\renewcommand{\vec}{\boldsymbol}
\newcommand{\vtheta}{{\vec{\theta}}}
\newcommand{\veta}{{\vec{\eta}}}
\newcommand{\vomega}{{\vec{\omega}}}

\newcommand{\vf}{{\vec{f}}}
\newcommand{\vg}{{\vec{g}}}

\newcommand{\word}[1]{{\em #1}}
\newcommand{\software}[1]{{\sc #1}}
\newcommand{\arrow}[1]{$\xmapsto{\text{#1}}$}

\begin{document}
\maketitle
\begin{abstract}
  Morphological segmentation has traditionally been modeled with non-hierarchical models, which yield flat segmentations as output. In many cases, however, proper morphological analysis requires hierarchical structure---especially in the case of
  derivational morphology. In this work, we introduce a discriminative, joint model of morphological segmentation along with the
  orthographic changes that occur during word formation. To the best
  of our knowledge, this is the first attempt to approach
  discriminative segmentation with a context-free model. Additionally,
  we release an annotated treebank of 7454 English words with
  constituency parses, encouraging future research in this area.\footnote{\textit{We found post publication that CELEX \cite{baayen1993celex} has annotated words for hierarchical morphological segmentation as well.}}
\end{abstract}

\section{Introduction}\label{sec:introduction}
In NLP, supervised morphological segmentation has typically been
viewed as either a sequence-labeling or a segmentation task \cite{segstudy}. In
contrast, we consider a hierarchical approach,
employing a context-free grammar (CFG). CFGs provide a richer model of
morphology: They capture (i) the intuition that words themselves have
internal constituents, which belong to different categories, as well
as (ii) the order in which affixes are attached. 
Moreover, many
morphological processes, e.g., compounding and
reduplication, are best modeled as hierarchical; thus,
context-free models are expressively more appropriate.

The purpose of morphological segmentation is to decompose words into
smaller units, known as morphemes, which are typically taken to be the
smallest meaning-bearing units in language. This work concerns itself
with modeling hierarchical structure over these morphemes. Note a
simple flat morphological segmentation can also be straightforwardly
derived from the CFG parse tree. Segmentations have found use in a
diverse set of NLP applications, e.g., automatic speech recognition
\cite{afify2006use}, keyword spotting \cite{narasimhanmorphological},
machine translation \cite{clifton2011combining} and parsing
\cite{seeker2015graph}. In contrast to prior work, we focus on 
  \ouremph{canonical segmentation}, i.e., we seek to jointly model orthographic
changes and segmentation. For instance, the canonical segmentation of
\word{untestably} is
\word{un}$+$\word{test}$+$\word{able}$+$\word{ly}, where we map
\word{ably} to \word{able}$+$\word{ly}, restoring the letters
\word{le}.

We make two contributions: (i) We introduce a joint model for
canonical segmentation with a CFG backbone. We experimentally
show that this model outperforms a semi-Markov model on flat segmentation.
(ii) We release the
first morphology treebank,
consisting of 7454 English word types, each annotated with a full constituency
parse.

\begin{figure*}
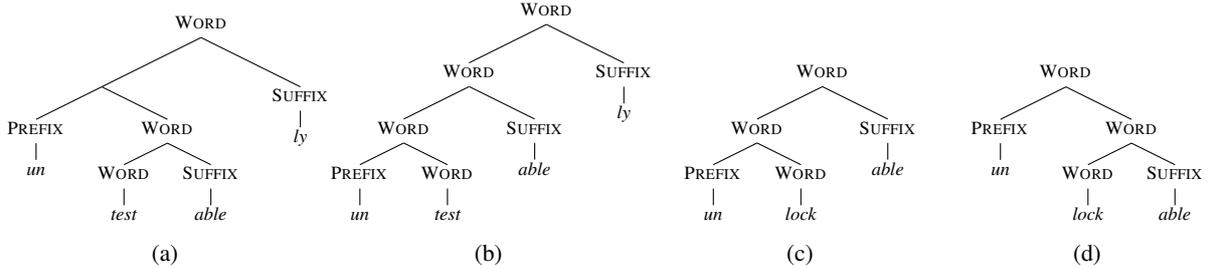

  \begin{subfigure}[b]{0.26 \textwidth}
    {\scriptsize \Tree [.{\sc Word} [[.{\sc Prefix} {\em un} ] [.{\sc Word} [.{\sc Word} {\em test} ] [.{\sc Suffix} {\em able} ]]] [.{\sc Suffix} {\em ly} ]]}
        \caption{}
  \end{subfigure}
  \begin{subfigure}[b]{0.26 \textwidth}
        {\scriptsize \Tree [.{\sc Word} [.{\sc Word} [.{\sc Word} [.{\sc Prefix} {\em un} ] [.{\sc Word} {\em test} ]] [.{\sc Suffix} {\em able} ]] [.{\sc Suffix} {\em ly} ]]}
        \caption{}
      \end{subfigure}
  \begin{subfigure}[b]{0.24\textwidth}
    {\scriptsize \Tree [.{\sc Word} [.{\sc Word} [.{\sc Prefix} {\em un} ] [.{\sc Word} {\em lock} ]] [.{\sc Suffix} {\em able} ]] }
    \caption{}
      \end{subfigure}
  \begin{subfigure}[b]{0.22\textwidth}
    {\scriptsize \Tree [.{\sc Word} [.{\sc Prefix} {\em un} ] [.{\sc Word} [.{\sc Word} {\em lock} ] [.{\sc Suffix} {\em able} ]]] }
    \caption{}
  \end{subfigure}
  \caption{Canonical segmentation parse trees for
    \word{untestably} and \word{unlockable}. For both words,
    the scope of \word{un} is ambiguous. Arguably, (a) is the only correct
    parse tree for \word{untestably}; the reading associated
    with (b) is hard to get. On the other hand,
    \word{unlockable} is truly ambiguous between
``able to be unlocked'' (c) and ``unable to be
locked'' (d).}
  \label{fig:trees}
\end{figure*}

\section{The Case For Hierarchical Structure}
Why should we analyze morphology hierarchically? It is true that we
can model much of morphology with finite-state machinery
\cite{beesley2003finite}, but there are, nevertheless, many cases
where hierarchical structure appears requisite. For instance, the flat
segmentation of the word
\word{untestably}$\mapsto$\word{un}$+$\word{test}$+$\word{able}$+$\word{ly}
is missing  important information about
how the word was  derived. The correct {\em parse}
[[\word{un}[[\word{test}]\word{able}]]\word{ly}], on the other hand,
does tell us that this is the order in which the complex form was
derived:
\begin{quote}
\word{test}\arrow{able}\word{testable}\arrow{un}\word{untestable}\arrow{ly}\word{untestably}.
\end{quote}
This gives us insight into the structure of the lexicon---we expect that the segment \word{testable} exists as
an independent word, but \word{ably} does not.

Moreover, a flat segmentation is often semantically ambiguous. 
There are two potentially valid readings of
\word{untestably} 
depending on how the negative prefix \word{un} scopes. The correct
tree (see \cref{fig:trees}) yields the reading ``in
the manner of not able to be tested.'' A second---likely infelicitous
reading---where the segment \word{untest} forms a constituent
yields the reading ``in a manner of being able to
untest.'' Recovering the hierarchical structure allows us to
select the correct reading; note there are even cases of
true ambiguity; e.g.,
\word{unlockable} has two
readings: ``unable to be locked'' and  ``able to
be unlocked.''
  

We also note that theoretical linguists often implicitly assume
a context-free treatment of word formation, e.g., by
employing brackets to indicate different levels of
affixation. Others have explicitly modeled word-internal
structure with grammars
\cite{selkirk0,marvin2002topics}.


 
\section{Parsing the Lexicon}\label{sec:model}
A novel component of this work is the development of a discriminative 
parser \cite{finkel2008efficient,hall2014less} for morphology. The
goal is to define a probability distribution over all trees
that could arise from the input word, after reversal of orthographic and
phonological processes. We employ the simple grammar shown
in \cref{tab:grammar}.
Despite its simplicity, it models the {\em order
in which  morphemes are attached}.\looseness=-1


More formally, our goal is to map a surface form $w$ (e.g.,
$w$$=$\word{untestably}) into its underlying canonical form $u$ (e.g.,
$u$$=$\word{untestablely}) and then into a parse tree $t$ over its
morphemes. We assume $u, w \in \Sigma^*$, for some discrete alphabet
$\Sigma$.\footnote{For efficiency, we  assume $u \in
  \Sigma^{|w|+k}$, $k=5$.} Note that a parse
tree over the string implicitly defines a flat segmentation given our
grammar---one can simply extract the characters spanned by all
preterminals in the resulting tree. Before describing the joint
model in detail, we first consider its pieces individually.\looseness=-1


\subsection{Restoring Orthographic Changes}
To extract a {\em canonical segmentation}
\cite{naradowsky2009improving,cotterell2016canonical}, we restore
orthographic changes that occur during word formation.
To this end,
we define the score function
\begin{equation}
\textit{score}_{\veta}(u, a,  w) = \exp\left(\vg(u,a, w)^{\top} \veta\right)
\end{equation}
where $a$ is a monotonic alignment between the strings $u$ and $w$. The goal is for $\text{score}_{\veta}$ to assign
higher values to better matched pairs, e.g., ($w$$=$\word{untestably},
$u$$=$\word{untestablely}).
We refer to \newcite{dreyer2008latent} for a  thorough exposition.

For ease of computation, we can encode this function as a weighted
finite-state machine (WFST) \cite{mohri2002weighted}. This requires,
however, that the feature function $\vg$ factors over the topology of
the finite-state encoding. Since our model conditions on the word $w$,
the feature function $\vg$ can extract features from {\em any} part of
this string. Features on the output string, $u$, however, are more
restricted. In this work, we employ a bigram model over output
characters. This implies that each state remembers exactly one
character: the previous one. See \newcite{cotterell-peng-eisner-2014}
for details.
We can compute the score for two strings $u$ and $w$ using a weighted generalization of the Levenshtein algorithm.
Computing the partition function requires a different dynamic
program, which runs in $\mathcal{O}(|w|^2\cdot|\Sigma|^2)$ time.
Note that since $|\Sigma| \approx
26$ (lower case English letters), it takes roughly $26^2 = 676$ times
longer to compute the partition function than to score a pair of
strings.\looseness=-1

Our model includes several simple feature templates, including
features that fire on individual edit actions as well as conjunctions
of edit actions and characters in the surrounding context. See
\newcite{cotterell2016canonical} for details.




\begin{table}
\centering
\begin{tabular}{lll}
  $\text{\sc Root}$ & $\rightarrow$ & $\text{\sc Word}$ \\
  $\text{\sc Word}$ & $\rightarrow$ & $\text{\sc Prefix }\text{\sc Word}$ \\
  $\text{\sc Word}$ & $\rightarrow$ & $\text{\sc Word }\text{\sc Suffix}$\\ 
  $\text{\sc Word}$  & $\rightarrow$ & $\Sigma^+$ \\ 
 $\text{\sc Prefix}$ & $\rightarrow$ & $\Sigma^+$ \\
  $\text{\sc Suffix}$ & $\rightarrow$ & $\Sigma^+$ 
\end{tabular}
\caption{The context-free grammar used in this work to model
  word formation. The productions closely resemble those of
\protect\newcite{johnson2006adaptor}'s 
Adaptor Grammar.}
\label{tab:grammar}
\end{table}

\subsection{Morphological Analysis as Parsing}
Next, we need to score an underlying canonical form (e.g.,
$u$$=$\word{untestablely}) together with a parse tree (e.g.,
$t$$=$[[\word{un}[[\word{test}]\word{able}]]\word{ly}]).  Thus, we
define the parser score with the following function
\begin{align}
  \textit{score}_{\vomega}(t, u) &= \exp\left(\sum_{\pi \in \Pi(t)}
  \vf(\pi, u)^{\top} \vomega\right)
\end{align}
where $\Pi(t)$ is the set of \emph{anchored productions} in the tree $t$.
An anchored production $\pi$ is a grammar rule in Chomsky
normal form attached to a span, e.g., $A_{i,k}
\rightarrow B_{i,j} C_{j,k}$. Each $\pi$ is then
assigned a weight by the linear function $\vf(\pi,
u)^{\top} \vomega$, where the function $\vf$ extracts relevant
features from the anchored production as well as the corresponding span
of the underlying form $u$. This model is typically
referred to as a weighted CFG (WCFG) \cite{smith2007weighted} or a CRF parser.



For $\vf$, we define three span features: (i)
indicator features on the span's segment, (ii) an indicator feature that
fires if the segment appears in an external corpus\footnote{We use the
 Wikipedia dump from 2016-05-01.} and (iii) the conjunction of
the segment with the label (e.g., {\sc Prefix}) of the subtree root. Following
\newcite{hall2014less}, we employ an indicator feature for each production as well as
production backoff features.

\reversemarginpar
\section{A Joint Model}
Our
complete model is a joint CRF \cite{koller2009probabilistic} where each of the above scores
are factors. We define the following probability distribution over trees, canonical forms
and their alignments to the original word
\begin{align}
  p_{\vtheta}(t, &a, u \mid w) = \\
  &\frac{1}{Z_{\vtheta}(w)}\, \textit{score}_{\vomega}(t, u) \cdot \textit{score}_{\veta}(u, a, w)\nonumber 
\end{align}
where $\vtheta = \{ \vomega, \veta\}$ is the parameter vector and the normalizing partition function as
\begin{align}
    &Z_{\vtheta}(w) =  \nonumber\\
    &\sum_{u' \in \Sigma^{|w|+k}} \sum_{a\in A(u', w)} \\ 
    &\hspace{1cm} \sum_{t' \in \mathcal{T}(u')} \nonumber \textit{score}_{\vomega}(t', u') \cdot \textit{score}_{\veta}(u', a,
   w) \nonumber
\end{align}
where $\mathcal{T}(u)$
is the set of all parse trees for the string $u$.
This involves a sum over all possible underlying orthographic forms and all parse trees
for those forms.\looseness=-1

The joint approach has the advantage that it allows both factors to work together to influence the choice of the underlying form $u$. This is useful as the parser
now has access to which words are attested in the
language; this helps guide the relatively weak transduction model. On
the downside, the partition function $Z_{\vtheta}$ now involves a
sum over all strings in $\Sigma^{|w|+k}$ {\em and} all
possible parses of each string! 
Finally, we define the marginal distribution over trees and underlying forms as 
\begin{equation}\label{eq:marginalized}
    p_{\vtheta}(t, u \mid w) = \!\!\!\sum_{a \in A(u, w)} \!\!\! p_{\vtheta}(t, a, u \mid w)
\end{equation}
where $A(u, w)$ is the set of all monotonic alignments between $u$ and $w$.
The marginalized form in \cref{eq:marginalized} is our final
model of morphological segmentation since we are not interested in the latent alignments $a$. 
\ryan{Rewrite this section.}


\begin{table*}
\centering
    \begin{tabular}{l  ccc  c} \toprule
      & \multicolumn{3}{c}{Segmentation} & Tree \\ \midrule
    & {\em Morph.} $F_1$  & {\em Edit} & {\em Acc.}  & {\em Const.} $F_1$ \\ \midrule
      {\em Flat} & 78.89 (0.9) & 0.72 (0.04) & 72.88 (1.21) & N/A \\ 
      {\em Hier} & {\bf 85.55} (0.6) & {\bf 0.55} (0.03) & {\bf 73.19} (1.09) & 79.01 (0.5) \\ \bottomrule
  \end{tabular} 
  \caption{Results for the 10 splits of the
    treebank. 
Segmentation quality is measured by
morpheme $F_1$, edit distance and accuracy;
tree quality by constituent $F_1$.}
  \label{tab:seg-results}
\end{table*}

\subsection{Learning and Inference}
We use stochastic gradient descent to
optimize the
log-probability of the training data $\sum_{n=1}^N \log
p_{\vtheta}(t^{(n)}, u^{(n)} \mid w^{(n)})$; this requires the
computation of the gradient of the partition function
$\nabla_{\vtheta}\log Z_{\vtheta}$.
We may view this gradient as an expectation:
\begin{align}
\label{eq:log-partition}
  &\nabla_{\vtheta} \log Z_{\vtheta}(w)= \\ 
&\hspace{0.25cm}\mathbb{E}_{(t, a, u)\sim p_{\vtheta}(\cdot \mid w)} 
    \left(\sum_{\pi \in \Pi(t)} {\boldsymbol f}(\pi, u)^{\top} + {\boldsymbol g}(u, a, w)^{\top} \right) \nonumber
\end{align}
We provide the full derivation in \cref{app:derivation} with an additional Rao-Blackwellization step that we make use of in the implementation.
While the sum over
all underlying forms and trees in \cref{eq:log-partition} may be achieved in polynomial time (using the Bar-Hillel construction), we make use of an importance-sampling estimator, derived by \newcite{cotterell2016canonical}, which is faster in practice.
Roughly speaking, we approximate the
hard-to-sample-from distribution $p_{\vtheta}$ by taking samples from
an easy-to-sample-from proposal distribution $q$. Specifically, we employ a pipeline model for $q$ consisting of WFST and then a 
WCFG sampled from consecutively. We then {\em reweight} the samples using the {\em unnormalized} score from $p_{\vtheta}$.
Importance sampling has found many uses in NLP ranging from language modeling \cite{bengio2003quick}
and neural MT \cite{JeanCMB15}
to parsing
\cite{dyer2016recurrent}.
Due to a lack of space, we omit the
derivation of the importance-sampled approximate gradient.\looseness=-1

\subsection{Decoding} We also decode by importance sampling. Given
$w$, we sample canonical forms $u$ and then run the CKY algorithm
to get the highest scoring tree.

\section{Related Work}\label{sec:prev-work}
We believe our attempt to train discriminative grammars for morphology
is novel. Nevertheless, other researchers have described parsers for
morphology. Most of this work is unsupervised:
\newcite{johnson2007bayesian} applied a Bayesian PCFG to unsupervised
morphological segmentation.  Similarly, Adaptor Grammars
\cite{johnson2006adaptor}, a non-parametric Bayesian generalization of
PCFGs, have been applied to the unsupervised version of the task
\cite{botha2013adaptor,sirts2013minimally}. Relatedly, \newcite{schmid2005disambiguation} performed unsupervised
disambiguation of a German morphological analyzer \cite{schmid2004smor}
using a PCFG, using the inside-outside algorithm \cite{baker1979trainable}. Also, discriminative parsing approaches have been applied to the related
problem of Chinese word segmentation \cite{zhang2014character}.

\section{Morphological Treebank}\label{sec:tree-bank}
Supervised morphological segmentation has historically been treated as a
segmentation problem, devoid of hierarchical structure. A core reason
behind this is that---to the best of our knowledge---there are no
hierarchically annotated corpora for the task. To remedy
this, we provide
tree annotations for a subset of the English portion of
CELEX \cite{baayen1993celex}. We reannotated 7454 English types with a
full constituency parse.\footnote{In many cases, we corrected the flat segmentation as well.} The resource will be freely available for future
research.

\subsection{Annotation Guidelines}
The annotation of the morphology treebank was guided by three core principles.
The first principle concerns \ouremph{productivity}:
we  exclusively annotate {\em productive} morphology.
In the context of morphology,
productivity refers to the degree that native speakers actively employ the
affix to create new words \cite{aronoff1976word}. We believe
that for NLP
applications, we should focus on productive affixation. Indeed,
this sets our corpus apart from many existing morphologically
annotated corpora such as CELEX. For example, CELEX contains
\word{warmth}$\mapsto$\word{warm$+$th}, but 
\word{th} is not a productive suffix and cannot be used to create new
words. Thus, we {\em do not} want to analyze
\word{hearth}$\mapsto$\word{hear$+$th} or, in general, allow
\word{wug}$\mapsto$\word{wug$+$th}. Second, we annotate for 
  \ouremph{semantic coherence}.  When there are several candidate parses, we
choose the one that is best compatible with the compositional
semantics of the derived form.\looseness=-1

Interestingly, multiple trees can be considered valid depending on the
linguistic tier of interest. Consider the word \word{unhappier}.  From
a semantic perspective, we have the parse [[\word{un} [\word{happy}]]
  \word{er}] which gives us the correct meaning ``not happy to a
greater degree.''  However, since the suffix \word{er} only attaches
to mono- and bisyllabic words, we get [\word{un}[[\word{happy}]
    \word{er}]] from a phonological perspective. In the linguistics
literature, this problem is known as the bracketing paradox
\cite{pesetsky1985morphology,embick2015morpheme}. We annotate
exclusively at the syntactic-semantic tier.\looseness=-1

Thirdly, in the context of derivational morphology, we {\em force
  spans to be words} themselves. Since derivational morphology---by
definition---forms new words from existing words
\cite{lieber2014oxford}, it follows that each span rooted with {\sc
  Word} or {\sc Root} in the correct parse corresponds to a word in
the lexicon. For example, consider \word{unlickable}.  The correct
parse, under our scheme, is [\word{un} [[\word{lick}] \word{able}]].
Each of the spans (\word{lick}, \word{lickable} and \word{unlickable})
exists as a word. By contrast, the parse [[\word{un} [\word{lick}]]
  \word{able}] contains the span \word{unlick}, which is not a word
in the lexicon. The span in the segmented form may involve changes, e.g., [\word{un} [[\word{achieve}]
    \word{able}]], where \word{achieveable} is not a word, but
\word{achievable} (after deleting \word{e}) is.


\section{Experiments}\label{sec:experiments}
We run a simple experiment to show the empirical utility of parsing
words---we compare a WCFG-based canonical segmenter with the
semi-Markov segmenter introduced in \newcite{cotterell2016canonical}.
We divide the corpus into 10 distinct train/dev/test splits with 5454
words for train and 1000 for each of dev and test. We report three
evaluation metrics: full form accuracy, morpheme $F_1$
\cite{van1999memory} and average edit distance to the gold
segmentation with boundaries marked by a distinguished symbol.  For
the WCFG model, we also report constituent $F_1$---typical for
sentential constituency parsing--- as a baseline for future
systems. This $F_1$ measures how well we predict the whole tree (not just a segmentation).  For all models, we use $L_2$
regularization and run 100 epochs of \software{AdaGrad}
\cite{duchi2011adaptive} with early stopping.
We tune the regularization coefficient by grid search
  considering $\lambda \in \{0.0, 0.1, 0.2, 0.3, 0.4, 0.5\}$.

\subsection{Results and Discussion}
\cref{tab:seg-results} shows the results.
The hierarchical WCFG model outperforms the flat semi-Markov
model on all metrics on the segmentation task. This shows that
modeling structure among the morphemes, indeed, does help
segmentation. The largest improvements are found under the morpheme
$F_1$ metric ($\approx6.5$ points). In contrast, accuracy
improves by $<1\%$. Edit distance is in between with an improvement of $0.2$
characters. Accuracy, in general, is an all or nothing metric since it
requires getting every canonical segment correct. Morpheme $F_1$, on the
other hand, gives us partial credit. Thus, what this shows us is that
the WCFG gets a lot more of the morphemes in the held-out set correct,
even if it only gets a few more complete forms correct. We provide
additional results evaluating the entire tree with constituency $F_1$
as a future baseline.



\section{Conclusion}\label{sec:conclusion}
We presented a discriminative CFG-based model for canonical
morphological segmentation and showed empirical improvements on its
ability to segment words under three metrics. We argue that our
hierarchical approach to modeling morphemes is more often appropriate
than the traditional flat segmentation.
Additionally, we have annotated 7454 words with a morphological
constituency parse. The corpus is available online at {\small \tt
  \url{http://ryancotterell.github.io/data/ morphological-treebank}} to
allow for exact comparison and to spark future research.

\section*{Acknowledgements}
The first author was supported by a DAAD Long-Term Research
Grant and an NDSEG fellowship. 
The third author was supported by DFG (SCHU 2246/10-1).

\bibliographystyle{acl_natbib}
\bibliography{morpho-seg-inside-out}

\onecolumn
\appendix
\section{Derivation of Eq.~\ref{eq:log-partition}}\label{app:derivation}
Here we provide the gradient of the log-partition function as an expectation:
\begin{align}
\nabla_{\vtheta}& \log Z_{\vtheta}(w) = \frac{1}{Z_{\vtheta}(w)} \nabla_{\vtheta} Z_{\vtheta}(w) \hspace{5cm} \\
&= \frac{1}{Z_{\vtheta}(w)} \nabla_{\vtheta} \left(\sum_{u' \in \Sigma^{|w|+k}} \sum_{a\in A(u', w)} \sum_{t' \in \mathcal{T}(u')} \textit{score}_{\vomega}(t', u') \cdot \textit{score}_{\veta}(u', a,
   w) \right) \nonumber \\
  &= \frac{1}{Z_{\vtheta}(w)} \sum_{u' \in \Sigma^{|w|+k}} \sum_{a\in A(u', w)} \sum_{t' \in \mathcal{T}(u')} \nabla_{\vtheta} \left(\textit{score}_{\vomega}(t', u') \cdot \textit{score}_{\veta}(u', a,
   w)  \right) \nonumber \\
   &= \frac{1}{Z_{\vtheta}(w)} \sum_{u' \in \Sigma^{|w|+k}} \sum_{a\in A(u', w)} \sum_{t' \in \mathcal{T}(u')}  \Big( \textit{score}_{\veta}(u', a,
   w) \cdot \nabla_{\vomega} \textit{score}_{\vomega}(t', u') \nonumber \\ 
   &\hspace{3.0cm} + \textit{score}_{\vomega}(t', u') \cdot \nabla_{\veta} \textit{score}_{\veta}(u', a,
   w)  \Big) \nonumber \\
      &= \frac{1}{Z_{\vtheta}(w)} \sum_{u' \in \Sigma^{|w|+k}} \sum_{a\in A(u', w)} \sum_{t' \in \mathcal{T}(u')} \textit{score}_{\veta}(u', a,
   w) \cdot \textit{score}_{\vomega}(t', u')  \left(\sum_{\pi \in \Pi(t')} {\boldsymbol f}(\pi, u')^{\top} + {\boldsymbol g}(u', a, w)^{\top} \right)  \nonumber \\
    &= \sum_{u' \in \Sigma^{|w|+k}} \sum_{a\in A(u', w)} \sum_{t' \in \mathcal{T}(u')} \frac{\textit{score}_{\veta}(u', a,
   w) \cdot \textit{score}_{\vomega}(t', u')}{Z_{\vtheta}(w)} \left(\sum_{\pi \in \Pi(t')} {\boldsymbol f}(\pi, u)^{\top} + {\boldsymbol g}(u', a, w)^{\top} \right) \nonumber \\
       &= \mathbb{E}_{(t, a, u)\sim p_{\vtheta}(\cdot \mid w)}  \left(\sum_{\pi \in \Pi(t)} {\boldsymbol f}(\pi, u)^{\top} + {\boldsymbol g}(u, a, w)^{\top} \right)  \label{eq:high-variance}
\end{align}
The result above can be further improved through Rao-Blackwellization. 
In this case, when we sample a tree--underlying form pair $(t, u)$, we marginalize
out all alignments that could have given rise to the sampled pair. 
The final derivation is show below:
\begin{align}\label{eq:rb}
   \nabla_{\vtheta} \log Z_{\vtheta}(w)  &= \mathbb{E}_{(t, a, u)\sim p_{\vtheta}(\cdot \mid w)}  \left(\sum_{\pi \in \Pi(t)} {\boldsymbol f}(\pi, u)^{\top} + {\boldsymbol g}(u, a, w)^{\top} \right)  \nonumber  \\
   &=  \mathbb{E}_{(t, u)\sim p_{\vtheta}(\cdot \mid w)}  \left(\sum_{\pi \in \Pi(t)} {\boldsymbol f}(\pi, u)^{\top} + \sum_{a \in A(u, w)} p_{\vtheta}(a \mid u, w) {\boldsymbol g}(u, a, w)^{\top} \right)
\end{align}
This estimator in \cref{eq:rb} will have lower variance than \cref{eq:high-variance}.


\end{document}